\documentclass[runningheads]{llncs}
\usepackage[T1]{fontenc}
\usepackage{graphicx,verbatim}
\usepackage{hyperref}
 
\usepackage{url}
\usepackage{xcolor} \usepackage{booktabs}
\usepackage{rotating}
\usepackage{array}
\newcolumntype{C}[1]{>{\centering\arraybackslash}m{#1}}
\usepackage{multirow}
\usepackage{longtable}
\usepackage{adjustbox}  

\usepackage{pdflscape}  
\usepackage{rotating}   
\hypersetup{
    colorlinks=true,        
    linkcolor=blue,         
    citecolor=blue,    
    urlcolor=blue,         
    pdfborder={0 0 0}       
}
\usepackage{threeparttable}

\begin{document}
\title{KokushiMD-10: Benchmark for Evaluating Large Language Models on Ten Japanese National Healthcare Licensing Examinations}

\titlerunning{KokushiMD-10}
%

\author{
  Junyu Liu\inst{1}\thanks{These authors contributed equally.}\and
  Kaiqi Yan\inst{2}\protect\footnotemark[1]\and
  Tianyang Wang\inst{3}\protect\footnotemark[1]\and
  Qian Niu\inst{4}\and \\
  Momoko Nagai-Tanima\inst{1}\and
  Tomoki Aoyama\inst{1}
}

\authorrunning{J. Liu et al.}

\institute{%
  Kyoto University \and
  The Hong Kong University of Science and Technology \and
  Xi’an Jiaotong-Liverpool University \and
  The University of Tokyo \\
  \email{aoyama.tomoki.4e@kyoto-u.ac.jp}
}

\maketitle              
%
\begin{abstract}
Recent advances in large language models (LLMs) have demonstrated notable performance in medical licensing exams. However, comprehensive evaluation of LLMs across various healthcare roles, particularly in high-stakes clinical scenarios, remains a challenge. Existing benchmarks are typically text-based, English-centric, and focus primarily on medicines, which limits their ability to assess broader healthcare knowledge and multimodal reasoning. To address these gaps, we introduce KokushiMD-10, the first multimodal benchmark constructed from ten Japanese national healthcare licensing exams. This benchmark spans multiple fields, including Medicine, Dentistry, Nursing, Pharmacy, and allied health professions. It contains over 11588 real exam questions, incorporating clinical images and expert-annotated rationales to evaluate both textual and visual reasoning. We benchmark over 30 state-of-the-art LLMs, including GPT-4o, Claude 3.5, and Gemini, across both text and image-based settings. Despite promising results, no model consistently meets passing thresholds across domains, highlighting the ongoing challenges in medical AI. KokushiMD-10 provides a comprehensive and linguistically grounded resource for evaluating and advancing reasoning-centric medical AI across multilingual and multimodal clinical tasks.

\keywords{Reasoning Medical AI \and Multimodal Benchmark \and Japanese Licensing Examinations}
\end{abstract}


%
%
%

\section{Introduction}

Healthcare licensing examinations offer a rigorous and standardized framework for evaluating the clinical reasoning capabilities of large language models (LLMs) in realistic healthcare contexts \cite{morishita2024comparison}. These exams test not only factual recall but also complex diagnostic reasoning, making them valuable foundations for benchmark construction. For example, questions may involve interpreting laboratory results, imaging findings, or patient histories to arrive at differential diagnoses and appropriate treatment plans \cite{shieh2024assessing,al2025comparative}.  Accordingly, many benchmark datasets—such as MedQA \cite{jin2021disease}, MedMCQA \cite{pal2022medmcqa}, IgakuQA \cite{kasai2023evaluating}, and YakugakuQA \cite{sukeda2025japanese} have been developed from national licensing exams to support research in medical question answering and clinical decision-making. These datasets usually comprise real-exam multiple-choice questions spanning internal medicine, pharmacology, surgery, and public health. However, existing benchmarks often remain limited in scope, constrained by profession, modality, language, or depth of reasoning, which restricts their value for evaluating general-purpose medical AI systems across the full spectrum of healthcare domains.

While benchmarks such as MedMCQA \cite{pal2022medmcqa} and YakugakuQA \cite{sukeda2025japanese} have significantly expanded the scale and subject diversity of medical QA datasets, critical gaps remain. For example, MedMCQA spans 21 subjects from real-world Indian postgraduate medical entrance exams (AIIMS and NEET PG), yet focuses exclusively on Medicine-related content and excludes image-based questions, limiting its utility for multimodal assessment. Similarly, YakugakuQA is derived from Japan’s National Pharmacist Licensing Examinations and introduces profession-specific coverage beyond medicine. However, it only includes text-based questions and centers on pharmacy-related knowledge \cite{takagi2023performance}. Moreover, most existing benchmarks are either monolingual—typically in English \cite{liu2023benchmarking,xie2024medtrinity}—or narrowly target a single healthcare profession (e.g., dentists \cite{morishita2024comparison} or pharmacists \cite{sukeda2025japanese}), leaving important domains such as nursing, dentistry, and rehabilitation underrepresented. In addition, few benchmarks provide detailed reasoning annotations, such as chain-of-thought (CoT) explanations \cite{wei2022chain}, which are essential for investigating the underlying decision-making behavior of LLMs.


To address these limitations, we introduce KokushiMD-10\footnote{Benchmark will be available at: https://huggingface.co/datasets/collect/KokushiMD-10}, a large-scale multimodal benchmark constructed from ten Japanese national licensing examinations across diverse healthcare professions, including Medicine, Dentistry, Nursing, Physical Therapy, Occupational Therapy, Midwifery, Public Health Nursing, Radiologic Technology, Optometry, and Pharmacy. This broad coverage enables the evaluation of both domain-specific knowledge and cross-disciplinary generalization for LLMs. In addition to text-based questions, KokushiMD-10 includes a substantial number of image-based items (e.g., radiographs, clinical photographs), supporting robust assessment of visual reasoning in clinical contexts. All content is written in Japanese, addressing the linguistic gap in existing medical QA benchmarks. Furthermore, six of them are paired with detailed CoT explanations, enabling fine-grained analysis of model reasoning beyond mere accuracy. Our contributions are three-fold:


\begin{itemize}
\item To the best of our knowledge, we present the first benchmark spanning ten distinct healthcare professions, allowing a fine-grained evaluation of LLMs in both the core medical and allied health domains.
\item We release a large-scale QA dataset in Japanese that incorporates both textual and visual modalities, facilitating a comprehensive evaluation of vision-language understanding.
\item We benchmark a wide range of open and closed source LLMs on KokushiMD-10. The results show that even state-of-the-art models like GPT-4o \cite{chatgpt} passed 70\% exams while many models passed no exams, underscoring the challenges of safe and accurate medical reasoning.
\end{itemize}


\section{Methodology}

\subsection{Benchmark Overview}

KokushiMD-10 is a multimodal benchmark constructed from PDFs of official Japanese healthcare licensing examination issued by the Ministry of Health, Labor and Welfare between 2020 and 2024 \cite{mhlw2025goukaku}. The benchmark covers ten healthcare professions, including Medicine, Dentistry, Nursing, Pharmacy, Midwifery, Public Health Nursing, Physical Therapy, Occupational Therapy, Optometry, and Radiologic Technology, and 6 of them comprise both text-only and image-based multiple-choice questions. 

\begin{figure*}[t]
  \centering
  \includegraphics[width=\linewidth]{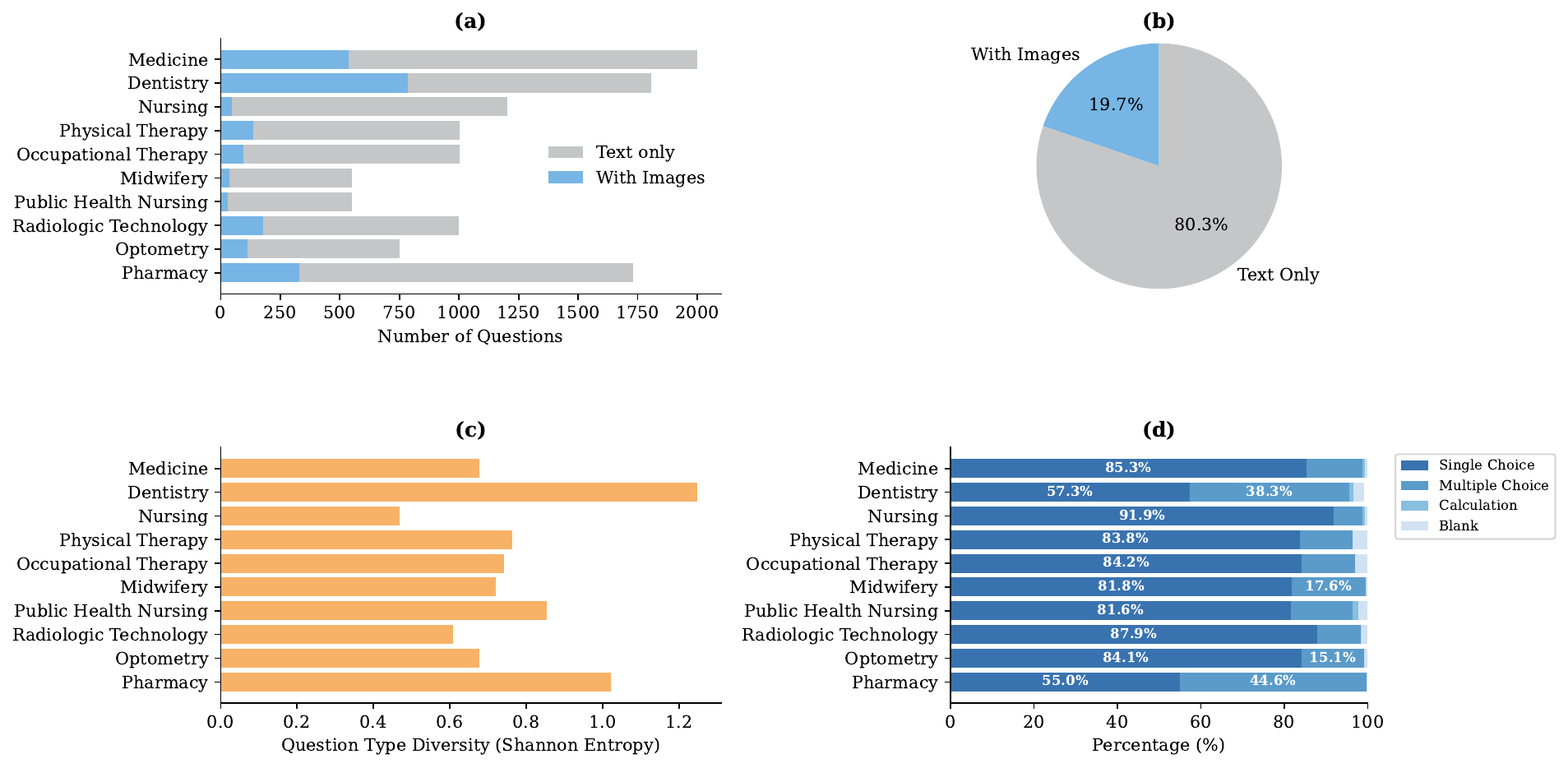}
  \caption{
    Dataset statistics across ten healthcare professions:
    (a) Total and image-based question counts per exam;
    (b) Overall proportion of image vs. text-only questions;
    (c) Question type diversity measured by Shannon entropy;
    (d) Distribution of question types (single-choice, multiple-choice, calculation, blank).
  }
  \label{fig:stats}
\end{figure*}

As shown in Fig.~\ref{fig:stats}, the dataset presents profession-level variation in question quantity, modality, and type. Image-based questions account for 19.7\% overall, with Dentistry showing the highest ratio. Medicine and Dentistry lead in total volume, while Nursing primarily features text-only items. Single-choice questions dominate in Medicine and Nursing, whereas Pharmacy and Dentistry contain more multiple-choice formats. Diversity in question types, quantified via Shannon entropy \cite{shannon1948}, is highest in Dentistry and lowest in Radiologic Technology.

\begin{figure}[t]
  \centering
  \includegraphics[width=\linewidth]{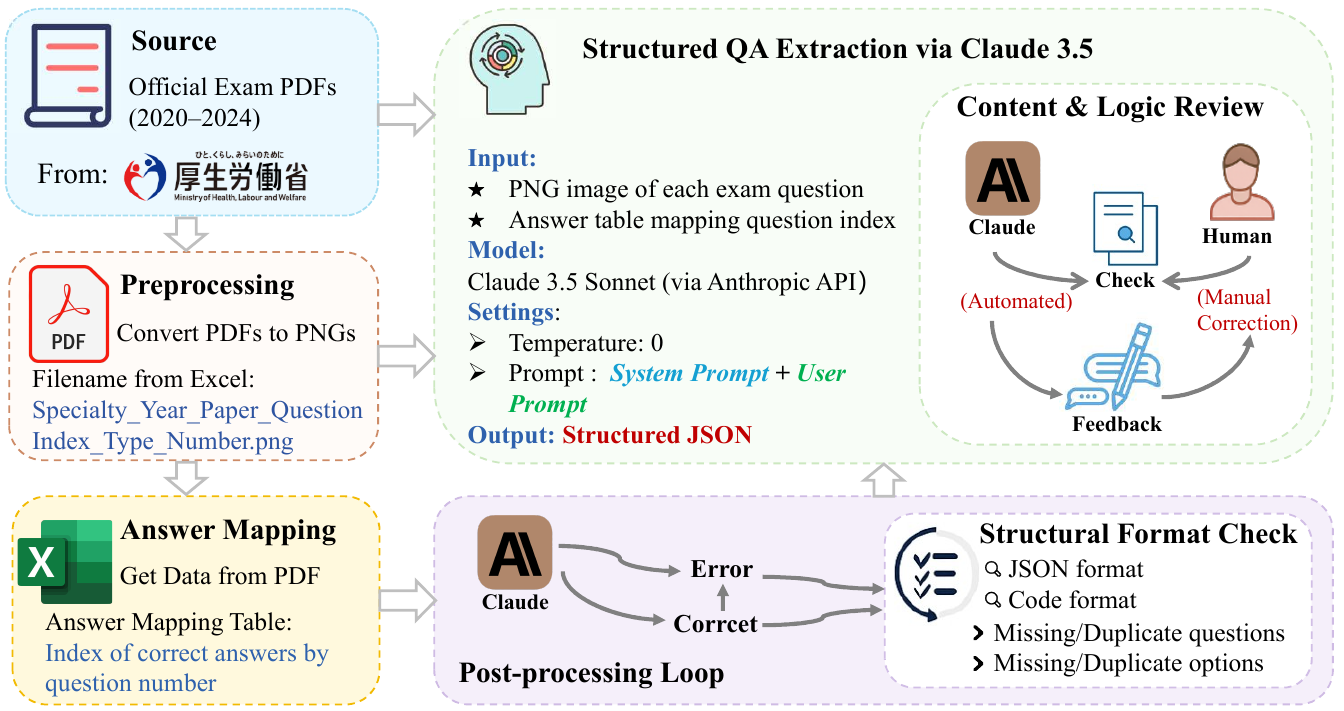}
  \caption{The construction pipeline of KokushiMD-10, including PDF-to-image preprocessing, structured QA extraction, answer alignment, human-in-the-loop correction, and structural format verification.}
  \label{fig:pipeline}
\end{figure}

\subsection{Benchmark Construction}
To ensure structural consistency and scalability, we employ a multi-stage pipeline (Fig.~\ref{fig:pipeline}) for benchmark construction.

\textbf{Data Processing and Answer Mapping.} Official PDF exam papers are first converted into high-resolution PNG images using Adobe Acrobat \cite{adobe_acrobat}. Each filename encodes key metadata such as the specialty, year, paper section, and question index in the format \texttt{Specialty\_Year\_Paper\_Index\_Type\_Number.png}. Answer mapping tables are extracted from the PDFs using Excel’s built-in table recognition and aligned with the question index to serve as the ground truth for downstream evaluation.

\textbf{Structured QA Extraction.} Each PNG question image is fed into Claude-3.5-Sonnet \cite{claude} via the Anthropic API using carefully engineered system-level and user-level prompts, with the temperature fixed at 0 to ensure deterministic outputs. The system prompt defines Claude as a data scientist responsible for organizing Japanese healthcare QA datasets, specifying a required JSON schema with fields for metadata (\texttt{year}, \texttt{section}, \texttt{index}), question components (\texttt{question}, \texttt{content}, options \texttt{a}--\texttt{e}), and format indicators (\texttt{answer}, \texttt{answer\_sub}, \texttt{text\_only}). The prompt includes comprehensive processing rules for sequential question handling, multimodal content detection (automatically setting \texttt{text\_only} to \texttt{false} when figures or tables are present), and quality control measures requiring cross-validation between question numbers and extracted indices. For each PNG input, the user prompt integrates metadata derived from the filename (specialty, exam year, question index) with task-specific instructions and references to the correct answer from our mapping table. The model generates structured JSON outputs capturing domain-specific subcategories (e.g., \texttt{corrected\_question\_index} for indexing consistency and \texttt{answer\_sub} for specialized fields like Pharmacy). This structured prompting strategy ensures that extracted QA entries are complete, valid, and semantically aligned with the original exam logic, even under varied question formats.

\textbf{Post-processing Loop.} Generated JSON files undergo a multi-stage validation procedure. Claude-3.5-Sonnet performs initial automated checks for structural format violations such as missing or duplicated entries. Scripts are made to check if there are missing or duplicated questions, images in the whole dataset. A team of human annotators then manually reviews selected samples for logical consistency, formatting issues, and semantic correctness. Feedback is iteratively incorporated through a looped refinement process. Special formatting, such as underlined text (\texttt{\textless u\textgreater ... \textless /u\textgreater}), chemical ions (e.g., HCO\textsubscript{3}\textsuperscript{--}), and isotope symbols (e.g., \textsuperscript{99m}Tc), is normalized to enhance parsing reliability for LLMs. This hybrid approach ensures the production of high-quality QA records aligned with medical-domain expectations. CoT explanations for six licensing examinations are constructed entirely from publicly available governmental and instructional sources \cite{mhlw_exams,ishiyaku_dental,tokyo_academy}. Additional exam preparation platforms were also referenced for explanatory consistency and structure \cite{guppy,mynavi_exam,contact_article}. English version of all the information in the benchmark is also provided.



\section{Experiments and Results}

\subsection{Baselines}

We adopt publicly available passing criteria from ten Japanese national healthcare licensing exams as reference points for human-level performance \cite{mhlw2025goukaku}. For example, the National Medical Licensing Examination in Japan requires candidates to achieve at least 80\% accuracy on designated essential questions and to meet the overall score thresholds. In addition, it includes "forbidden options" that, if selected, result in automatic failure—allowing us to benchmark high-risk decision errors. While some professions, such as dentistry and midwifery, apply composite scoring across general and scenario-based sections, others use varying sectional thresholds or qualitative grading. These diverse standards provide profession-specific baselines to assess model reliability in real-world clinical contexts.

\subsection{Implementation Details}

A total of 33 models were tested in text-only mode and 11 in multimodal mode on the KokushiMD-10 benchmark. Closed-source models from OpenAI, Anthropic, and Google Gemini \cite{gemini} were accessed through their public APIs, whereas each open-source model was executed from its Hugging Face release on a single compute node equipped with eight NVIDIA A100 GPUs.

Every query was issued with a two-part prompt. The system prompt, originally in Japanese, begins with the instruction, “As an expert [role], please answer the following [exam\_type] question.” It subsequently defines the required output format and directs the model to provide only that formatted answer. The user prompt contains the question text and, for multimodal runs, the associated images. All generated responses were recorded for subsequent evaluation.



\subsection{Evaluation Protocol and Metrics}

Questions are categorized into four types: \textit{Single Answer}, \textit{Multiple Choice}, \textit{Numerical Calculation}, and \textit{Blank}. Each type is scored with strict correctness rules—for example, multiple-choice questions are marked incorrect unless the selected options exactly match the correct set. Scenario-based questions in nursing and midwifery are evaluated with higher weight to reflect real-world decision-making. We report per-type accuracy, as well as domain-specific thresholds aligned with official exam standards. A small subset of blank or invalid questions is excluded following real-world practice. In total, we evaluate 33 models in text-only mode and 11 models in multimodal mode, across 50 licensing exams spanning ten professions and five years. A model is considered to have passed an exam if its score meets the official passing threshold for that exam. We record both the number of passed exams per model and the raw scores for each model in each exam\footnote{Scripts for evaluation will be available at: https://github.com/juniorliu95/KokushiMD-10}.

\subsection{Benchmark Performance}

 






\begin{table}[t]
\centering
\caption{Model pass/fail distribution across healthcare licensing exams.}
\label{tab:model_summary}
\begin{tabular}{lccc}
\toprule
Mode          & Passed          & Failed          & All-Zero \\
              & (out of total)  & All Exams       & Models   \\
\midrule
Text-only     & 12 / 33         & 21              & 6        \\
Multimodal    & 6 / 11          & 5               & 0        \\
\bottomrule
\end{tabular}
\end{table}

\begin{figure}[t]
    \centering
    \includegraphics[width=\linewidth]{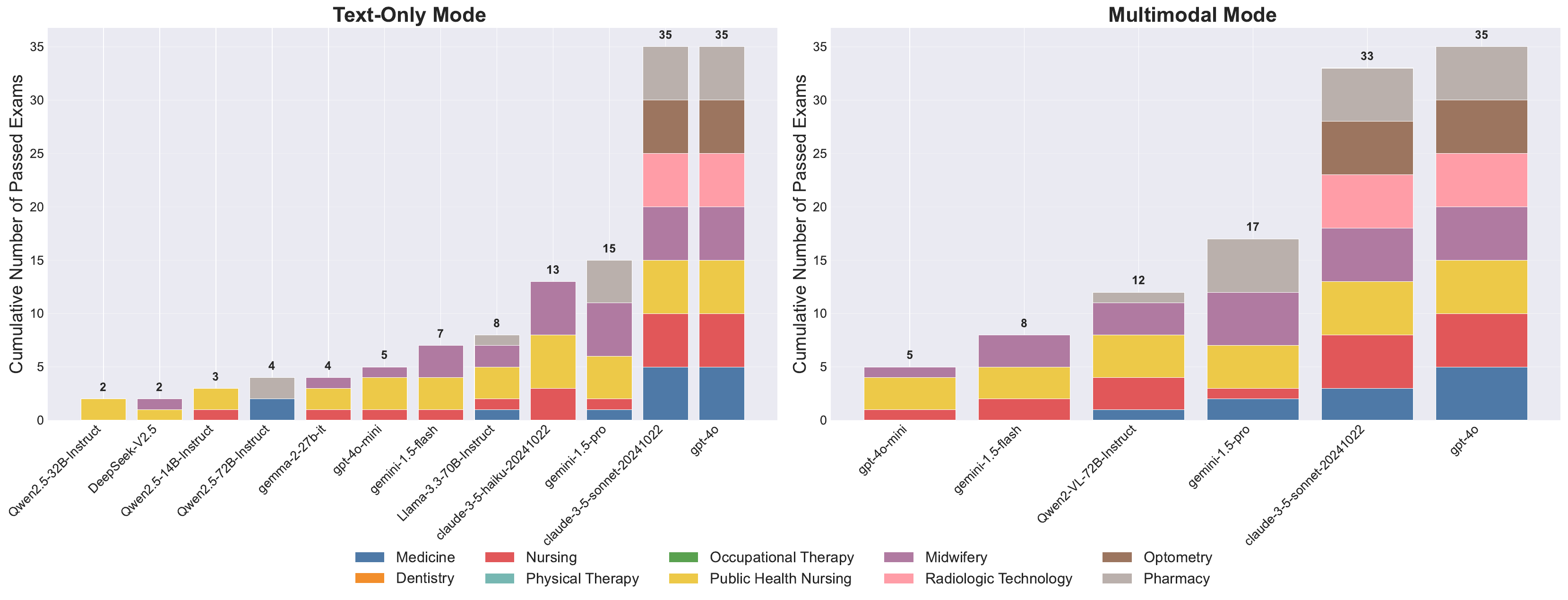}
\caption{Cumulative number of passed exams per model across 50 licensing exams (five years per healthcare profession). Colors indicate different healthcare subjects.}
\label{fig:50Q_2mode}
\end{figure}

\textbf{Pass/Fail Evaluation under Official Thresholds.}
We evaluate each model on 50 full licensing exams—five years of test data across ten healthcare professions—using official criteria for each. As shown in Table~\ref{tab:model_summary}, 33 models were tested in the Text-Only mode and 11 in the Multimodal mode. In the text-only setting, 12 models passed at least one exam, while 6 failed all exams with zero correct answers. In the multimodal setting, 6 models passed at least one exam, and all models scored non-zero. Fig.~\ref{fig:50Q_2mode} shows the cumulative number of passed exams per model, with colors indicating per-domain contributions. Only models with non-zero total scores are included.



\begin{table}[t]
\centering
\caption{Mean $\pm$ standard deviation of scores per healthcare specialty over five years (Text-Only Mode). The highest score in each subject is highlighted in bold.}
\label{tab:medical_performance}
\begin{threeparttable}
\adjustbox{width=\textwidth}{%
\begin{tabular}{|c|*{10}{C{2.2cm}|}}
\hline
\textbf{Model} & \textbf{Medicine} & \textbf{Dentistry} & \textbf{Nursing} & \textbf{Physical Therapy} & \textbf{Occupational Therapy} & \textbf{Public Health Nursing} & \textbf{Midwife} & \textbf{Radiological Technology} & \textbf{Optometry} & \textbf{Pharmacy} \\
\hline
Gemma-2-2B-IT & 0.00$\pm$0.00 & 0.00$\pm$0.00 & 0.20$\pm$0.40 & 0.00$\pm$0.00 & 0.40$\pm$0.49 & 0.00$\pm$0.00 & 0.00$\pm$0.00 & 0.00$\pm$0.00 & 0.20$\pm$0.40 & 0.00$\pm$0.00 \\
Phi-3.5-Vision-Instruct & 1.20$\pm$0.98 & 0.00$\pm$0.00 & 1.40$\pm$1.36 & 1.00$\pm$0.89 & 0.20$\pm$0.40 & 0.80$\pm$1.17 & 1.00$\pm$0.63 & 1.60$\pm$0.80 & 0.20$\pm$0.40 & 3.20$\pm$3.25 \\
QwQ-32B-Preview & 2.40$\pm$2.06 & 1.60$\pm$0.80 & 8.40$\pm$2.87 & 5.60$\pm$2.65 & 4.80$\pm$2.79 & 3.20$\pm$2.04 & 2.00$\pm$1.67 & 4.40$\pm$2.33 & 2.60$\pm$1.62 & 13.60$\pm$5.71 \\
GPT-3.5-Turbo & 11.40$\pm$3.44 & 5.00$\pm$2.61 & 0.80$\pm$0.75 & 2.20$\pm$2.04 & 2.60$\pm$1.74 & 1.40$\pm$1.50 & 0.40$\pm$0.49 & 0.20$\pm$0.40 & 0.00$\pm$0.00 & 14.00$\pm$5.06 \\
Qwen2.5-0.5B-Instruct & 22.00$\pm$3.69 & 3.80$\pm$2.99 & 8.40$\pm$7.00 & 0.80$\pm$1.17 & 3.00$\pm$1.79 & 3.60$\pm$2.24 & 2.20$\pm$2.99 & 0.00$\pm$0.00 & 0.00$\pm$0.00 & 4.00$\pm$2.83 \\
Llama-3.2-1B-Instruct & 27.40$\pm$6.53 & 20.80$\pm$7.28 & 17.60$\pm$3.44 & 8.40$\pm$5.08 & 10.60$\pm$3.77 & 8.80$\pm$2.93 & 6.00$\pm$2.68 & 3.40$\pm$1.62 & 6.20$\pm$1.83 & 17.20$\pm$5.31 \\
Qwen2.5-1.5B-Instruct & 69.00$\pm$8.81 & 3.60$\pm$5.82 & 2.20$\pm$3.49 & 0.40$\pm$0.80 & 0.40$\pm$0.80 & 0.00$\pm$0.00 & 0.00$\pm$0.00 & 0.00$\pm$0.00 & 2.20$\pm$2.48 & 12.80$\pm$11.57 \\
Llama-3.2-3B-Instruct & 142.00$\pm$10.66 & 119.20$\pm$5.98 & 88.40$\pm$12.77 & 46.80$\pm$7.14 & 48.00$\pm$3.85 & 34.20$\pm$6.31 & 36.40$\pm$6.47 & 29.60$\pm$4.32 & 25.40$\pm$6.09 & 123.60$\pm$5.71 \\
Phi-4 & 168.40$\pm$12.66 & 114.40$\pm$11.59 & 116.20$\pm$9.52 & 68.80$\pm$3.76 & 76.60$\pm$4.59 & 47.40$\pm$6.89 & 29.80$\pm$3.54 & 35.40$\pm$3.07 & 30.40$\pm$4.08 & 150.80$\pm$9.52 \\
Llama-3.1-8B-Instruct & 169.20$\pm$14.30 & 127.40$\pm$10.35 & 84.20$\pm$7.08 & 32.40$\pm$3.77 & 34.20$\pm$5.34 & 35.40$\pm$3.93 & 29.00$\pm$4.60 & 22.80$\pm$2.86 & 18.60$\pm$3.83 & 123.60$\pm$5.28 \\
Qwen2.5-3B-Instruct & 197.80$\pm$9.74 & 92.00$\pm$9.38 & 114.00$\pm$4.98 & 43.00$\pm$14.63 & 52.80$\pm$10.11 & 52.40$\pm$6.53 & 45.60$\pm$6.28 & 24.60$\pm$4.22 & 20.40$\pm$4.22 & 123.60$\pm$4.45 \\
Gemma-2-9B-IT & 255.60$\pm$11.11 & 188.20$\pm$18.68 & 169.20$\pm$13.56 & 91.00$\pm$9.08 & 99.40$\pm$14.01 & 68.40$\pm$7.71 & 65.20$\pm$2.79 & 51.20$\pm$7.52 & 44.60$\pm$4.50 & 261.20$\pm$12.43 \\
Claude-3.5-Haiku & 264.20$\pm$14.61 & 183.20$\pm$12.87 & 231.40$\pm$9.67 & 185.80$\pm$7.19 & 191.60$\pm$9.81 & 102.20$\pm$7.00 & 95.00$\pm$2.10 & 111.20$\pm$0.98 & 78.60$\pm$4.13 & 379.60$\pm$27.93 \\
Qwen2.5-7B-Instruct & 264.60$\pm$7.81 & 187.60$\pm$18.94 & 168.60$\pm$6.18 & 100.20$\pm$14.48 & 112.40$\pm$8.73 & 67.20$\pm$5.23 & 60.60$\pm$8.87 & 57.40$\pm$6.15 & 55.40$\pm$8.21 & 266.40$\pm$15.36 \\
Gemma-2-27B-IT & 311.40$\pm$12.04 & 227.00$\pm$26.33 & 206.00$\pm$9.17 & 130.40$\pm$5.46 & 141.20$\pm$9.68 & 85.60$\pm$9.91 & 77.60$\pm$5.57 & 73.40$\pm$8.66 & 57.00$\pm$6.96 & 349.20$\pm$14.23 \\
Qwen2.5-14B-Instruct & 314.20$\pm$8.52 & 232.80$\pm$25.81 & 193.60$\pm$11.57 & 121.20$\pm$8.75 & 128.60$\pm$14.64 & 81.00$\pm$11.78 & 73.00$\pm$6.90 & 68.40$\pm$7.63 & 65.60$\pm$3.44 & 340.80$\pm$19.17 \\
GPT-4o-Mini & 327.40$\pm$8.82 & 209.80$\pm$14.50 & 205.80$\pm$9.33 & 147.40$\pm$8.19 & 161.40$\pm$10.80 & 80.20$\pm$12.24 & 81.20$\pm$4.75 & 83.20$\pm$2.79 & 65.80$\pm$5.42 & 376.00$\pm$21.35 \\
DeepSeek-V2.5 & 335.40$\pm$11.74 & 230.00$\pm$10.00 & 197.60$\pm$13.50 & 138.60$\pm$8.16 & 151.20$\pm$9.30 & 77.00$\pm$8.51 & 74.40$\pm$7.66 & 94.40$\pm$7.26 & 68.20$\pm$6.55 & 351.60$\pm$7.31 \\
DeepSeek-V2.5-1210 & 344.80$\pm$8.28 & 235.40$\pm$19.74 & 192.00$\pm$10.45 & 134.80$\pm$7.08 & 141.40$\pm$9.48 & 71.80$\pm$6.62 & 69.20$\pm$10.24 & 91.00$\pm$5.76 & 63.20$\pm$7.36 & 360.00$\pm$12.84 \\
Gemini-1.5-Flash & 345.00$\pm$7.16 & 252.80$\pm$11.30 & 214.40$\pm$3.26 & 157.00$\pm$7.90 & 163.40$\pm$6.59 & 90.80$\pm$9.79 & 87.20$\pm$3.54 & 98.20$\pm$8.93 & 76.40$\pm$4.03 & 397.60$\pm$21.78 \\
Llama-3.1-70B-Instruct & 347.00$\pm$8.72 & 230.40$\pm$14.77 & 180.60$\pm$11.50 & 117.40$\pm$8.89 & 121.60$\pm$2.80 & 77.60$\pm$4.72 & 66.00$\pm$5.25 & 67.00$\pm$7.07 & 55.80$\pm$4.71 & 374.80$\pm$12.17 \\
Qwen2.5-32B-Instruct & 353.40$\pm$14.65 & 263.60$\pm$22.02 & 195.60$\pm$12.60 & 138.60$\pm$2.42 & 145.80$\pm$9.02 & 79.20$\pm$12.02 & 73.20$\pm$8.57 & 82.60$\pm$5.57 & 67.00$\pm$6.78 & 390.40$\pm$25.97 \\
Llama-3.3-70B-Instruct & 366.80$\pm$9.41 & 238.00$\pm$12.71 & 215.00$\pm$7.48 & 159.20$\pm$9.33 & 166.00$\pm$2.10 & 86.20$\pm$12.22 & 83.40$\pm$7.26 & 101.20$\pm$3.71 & 71.40$\pm$7.76 & 406.40$\pm$22.18 \\
Gemini-1.5-Pro & 384.60$\pm$11.74 & 301.60$\pm$16.72 & 221.20$\pm$8.47 & 177.00$\pm$8.92 & 182.20$\pm$8.13 & 96.20$\pm$9.68 & 95.00$\pm$4.60 & 108.60$\pm$7.00 & 90.40$\pm$3.98 & 465.20$\pm$27.03 \\
Qwen2.5-72B-Instruct & 385.40$\pm$5.89 & 274.20$\pm$24.10 & 187.20$\pm$13.26 & 140.20$\pm$11.79 & 146.00$\pm$12.54 & 71.60$\pm$7.86 & 74.60$\pm$3.83 & 84.60$\pm$7.47 & 66.80$\pm$5.04 & 422.80$\pm$14.46 \\
\textbf{claude-3-5-sonnet} & 413.80$\pm$7.39 & 315.60$\pm$16.02 & \textbf{266.60$\pm$2.94} & 223.00$\pm$8.32 & 226.80$\pm$5.53 & \textbf{124.20$\pm$7.25} & \textbf{116.20$\pm$3.54} & \textbf{147.20$\pm$3.12} & 111.00$\pm$4.24 & \textbf{536.80$\pm$18.70} \\
\textbf{GPT-4o} & \textbf{433.40$\pm$3.50} & \textbf{327.20$\pm$17.43} & 263.60$\pm$5.54 & \textbf{226.20$\pm$6.31} & \textbf{229.20$\pm$9.74} & 109.40$\pm$7.06 & 112.40$\pm$4.03 & 141.00$\pm$3.95 & \textbf{115.20$\pm$5.81} & 508.40$\pm$11.41 \\
\hline
\end{tabular}}
\begin{tablenotes}[flushleft]
\scriptsize
\item \raisebox{0.2ex}{\scalebox{0.6}{$\bullet$}}~Six zero-score models were excluded, leaving 27 text-only models.
\end{tablenotes}
\end{threeparttable}
\end{table}

\begin{table}[t]
\centering
\caption{Mean $\pm$ standard deviation of scores per healthcare specialty over five years (Multimodal Mode). The highest score in each subject is highlighted in bold.}
\label{tab:multimodal_exam_scores}
\begin{threeparttable}
\adjustbox{width=\textwidth}{%
\begin{tabular}{|c|*{10}{C{2.2cm}|}}
\hline
\textbf{Model} & \textbf{Medicine} & \textbf{Dentistry} & \textbf{Nursing} & \textbf{Physical Therapy} & \textbf{Occupational Therapy} & \textbf{Public Health Nursing} & \textbf{Midwife} & \textbf{Radiological Technology} & \textbf{Optometry} & \textbf{Pharmacy} \\
\hline
QVQ-72B-preview & 0.40$\pm$0.49 & 0.20$\pm$0.40 & 0.00$\pm$0.00 & 0.40$\pm$0.49 & 0.00$\pm$0.00 & 0.00$\pm$0.00 & 0.40$\pm$0.80 & 0.00$\pm$0.00 & 0.00$\pm$0.00 & 0.80$\pm$0.98 \\
Phi-3.5-vision-instruct & 2.00$\pm$1.10 & 1.60$\pm$1.96 & 1.80$\pm$2.14 & 1.00$\pm$0.89 & 0.20$\pm$0.40 & 1.20$\pm$1.17 & 1.00$\pm$0.63 & 2.60$\pm$1.36 & 2.20$\pm$2.04 & 13.20$\pm$5.46 \\
Qwen2-VL-2B-Instruct & 85.00$\pm$6.29 & 67.80$\pm$6.05 & 68.60$\pm$6.59 & 33.40$\pm$5.71 & 37.60$\pm$3.98 & 28.20$\pm$5.91 & 29.60$\pm$3.67 & 18.40$\pm$4.32 & 21.00$\pm$2.53 & 56.40$\pm$6.25 \\
Pixtral-12B-2409 & 168.00$\pm$20.65 & 122.80$\pm$15.51 & 105.60$\pm$9.89 & 37.20$\pm$5.19 & 42.40$\pm$5.75 & 45.20$\pm$4.58 & 39.60$\pm$10.61 & 31.20$\pm$2.64 & 25.00$\pm$3.46 & 139.60$\pm$10.31 \\
Qwen2-VL-7B-Instruct & 217.20$\pm$16.99 & 140.40$\pm$4.03 & 162.00$\pm$14.71 & 88.00$\pm$6.20 & 100.00$\pm$3.41 & 66.40$\pm$9.46 & 59.20$\pm$7.57 & 58.60$\pm$4.54 & 42.00$\pm$3.03 & 189.20$\pm$8.82 \\
gpt-4o-mini & 326.20$\pm$5.15 & 210.00$\pm$14.79 & 207.00$\pm$8.65 & 149.60$\pm$7.55 & 164.00$\pm$10.86 & 81.00$\pm$11.88 & 81.40$\pm$7.00 & 86.60$\pm$3.38 & 65.20$\pm$2.32 & 382.80$\pm$20.73 \\
gemini-1.5-flash & 340.60$\pm$4.27 & 251.60$\pm$27.95 & 215.20$\pm$3.19 & 157.80$\pm$7.88 & 163.80$\pm$7.41 & 91.40$\pm$9.56 & 87.40$\pm$4.54 & 97.40$\pm$6.50 & 74.80$\pm$5.15 & 401.60$\pm$19.41 \\
Qwen2-VL-72B-Instruct & 371.40$\pm$10.37 & 242.00$\pm$28.37 & 231.80$\pm$9.68 & 181.20$\pm$7.14 & 182.20$\pm$4.96 & 98.00$\pm$9.88 & 92.80$\pm$7.86 & 112.00$\pm$7.04 & 86.40$\pm$9.13 & 426.80$\pm$14.89 \\
gemini-1.5-pro & 383.60$\pm$13.37 & 296.40$\pm$17.41 & 223.00$\pm$8.27 & 177.20$\pm$8.98 & 182.00$\pm$9.70 & 97.40$\pm$10.31 & 96.40$\pm$5.16 & 111.80$\pm$6.85 & 89.00$\pm$6.23 & 478.80$\pm$24.55 \\
\textbf{claude-3-5-sonnet} & 389.80$\pm$13.98 & 297.40$\pm$26.51 & \textbf{268.40$\pm$2.42} & 223.60$\pm$7.61 & 228.80$\pm$5.64 & \textbf{125.80$\pm$5.64} & \textbf{116.20$\pm$3.43} & \textbf{150.20$\pm$4.66} & 110.40$\pm$5.31 & \textbf{551.20$\pm$18.96} \\
\textbf{gpt-4o} & \textbf{437.20$\pm$5.84} & \textbf{329.40$\pm$14.62} & 265.00$\pm$5.25 & \textbf{227.40$\pm$7.06} & \textbf{231.80$\pm$11.23} & 112.80$\pm$6.40 & 112.80$\pm$4.31 & 145.60$\pm$4.18 & \textbf{114.60$\pm$6.15} & 526.80$\pm$7.96 \\
\hline
\end{tabular}}
\end{threeparttable}
\end{table}

\noindent\textbf{Quantitative Score Analysis per Exam.}  Table~\ref{tab:medical_performance} reports scores across ten healthcare specialties in text-only mode. Among 27 models with non-zero scores, GPT-4o leads in four specialties (Medicine, Physical Therapy, Occupational Therapy, Optometry), while Claude-3.5-Sonnet dominates five areas (Nursing, Public Health Nursing, Midwife, Radiological Technology, Pharmacy). Notably, Pharmacy examinations yield the highest absolute scores (Claude: 536.8$\pm$18.70 vs GPT-4o: 508.4$\pm$11.41), whereas Dentistry consistently shows the lowest performance across all models. Gemini-1.5-Pro maintains competitive scores but achieves no category leadership.

The multimodal setting (Table~\ref{tab:multimodal_exam_scores}) preserves the same performance hierarchy with minimal score variations. GPT-4o and Claude-3.5-Sonnet maintain their respective domain leadership, with Claude showing slight improvements in Pharmacy (551.2$\pm$18.96) and Public Health Nursing categories. Cross-modal consistency indicates that visual information provides limited additional benefit for healthcare licensing performance, as rank ordering remains stable between text-only and multimodal evaluations across all tested models.

\section{Conclusion}

In this work, we introduced KokushiMD-10, a comprehensive multimodal benchmark constructed from ten Japanese national healthcare licensing examinations. The dataset spans a wide range of healthcare professions and includes both textual and visual questions, enabling thorough evaluation of LLMs across multiple clinical domains. Our evaluation of over 30 state-of-the-art models shows that, while some models perform reasonably well in isolated tasks, none consistently achieve the performance required for real-world clinical use, particularly in visual reasoning and complex decision-making. These findings highlight the need for more robust, generalizable, and multimodal medical AI systems. KokushiMD-10 lays the groundwork for their development by offering a rigorous and diverse evaluation framework. Future efforts will focus on strengthening clinical alignment and expanding real-world applicability.

\bibliographystyle{splncs04}
\bibliography{reference}

\begin{thebibliography}{10}
\providecommand{\url}[1]{\texttt{#1}}
\providecommand{\urlprefix}{URL }
\providecommand{\doi}[1]{https://doi.org/#1}

\bibitem{tokyo_academy}
Academy, T.: Qualification exam information center. \url{https://www.tokyo-ac.jp/qualification/info-exam/}, accessed: 2025-06-05

\bibitem{adobe_acrobat}
{Adobe Inc.}: Adobe acrobat pro. Computer software (2024), \url{https://acrobat.adobe.com}, version 2024.004.2054, accessed 8 Jun 2025

\bibitem{al2025comparative}
Al-Khater, K.M.K.: Comparative assessment of three ai platforms in answering usmle step 1 anatomy questions or identifying anatomical structures on radiographs. Clinical Anatomy  \textbf{38}(2),  186--199 (2025)

\bibitem{claude}
{Anthropic}: Claude 3 (opus release) (2025), \url{https://www.anthropic.com/claude}, accessed 8 Jun 2025

\bibitem{ishiyaku_dental}
Dental, I.: Overview and trends in dental national examinations. \url{https://www.ishiyaku-dental.jp/}, accessed: 2025-06-05

\bibitem{gemini}
{Google DeepMind}: Gemini 1.5 pro (2025), \url{https://gemini.google.com}, accessed 8 Jun 2025

\bibitem{mynavi_exam}
Healthcare, M.: National exam strategy and career information. \url{https://nurse.mynavi.jp/conts/kokushi_kouryaku/}, accessed: 2025-06-05

\bibitem{jin2021disease}
Jin, D., Pan, E., Oufattole, N., Weng, W.H., Fang, H., Szolovits, P.: What disease does this patient have? a large-scale open domain question answering dataset from medical exams. Applied Sciences  \textbf{11}(14), ~6421 (2021)

\bibitem{kasai2023evaluating}
Kasai, J., Kasai, Y., Sakaguchi, K., Yamada, Y., Radev, D.: Evaluating gpt-4 and chatgpt on japanese medical licensing examinations. arXiv preprint arXiv:2303.18027  (2023)

\bibitem{liu2023benchmarking}
Liu, J., Zhou, P., Hua, Y., Chong, D., Tian, Z., Liu, A., Wang, H., You, C., Guo, Z., Zhu, L., et~al.: Benchmarking large language models on cmexam-a comprehensive chinese medical exam dataset. Advances in Neural Information Processing Systems  \textbf{36},  52430--52452 (2023)

\bibitem{contact_article}
Media, C.: Overview of radiologic technologist national examination. \url{https://contact.ne.jp/media/?p=170}, accessed: 2025-06-05

\bibitem{mhlw2025goukaku}
{Ministry of Health, Labour and Welfare}: Qualification exam results announcement (2025), \url{https://www.mhlw.go.jp/kouseiroudoushou/shikaku_shiken/goukaku.html}, accessed 2025-06-07

\bibitem{mhlw_exams}
{Ministry of Health, Labour and Welfare of Japan}: National licensing examination overview and results. \url{https://www.mhlw.go.jp/kouseiroudoushou/shikaku_shiken/} (2024), accessed: 2025-06-05

\bibitem{morishita2024comparison}
Morishita, M., Fukuda, H., Muraoka, K., Nakamura, T., Yoshioka, I., Ono, K., Awano, S.: Comparison of the performance on the japanese national dental examination using gpt-3.5 and gpt-4. JJDEA  \textbf{40},  3--10 (2024)

\bibitem{chatgpt}
{OpenAI}: {ChatGPT} (gpt-4o model) (2025), \url{https://openai.com/chatgpt}, accessed 8 Jun 2025

\bibitem{pal2022medmcqa}
Pal, A., Umapathi, L.K., Sankarasubbu, M.: Medmcqa: A large-scale multi-subject multi-choice dataset for medical domain question answering. In: Conference on health, inference, and learning. pp. 248--260. PMLR (2022)

\bibitem{guppy}
Platform, G.J.: National exam overview for public health roles. \url{https://www.guppy.jp/}, accessed: 2025-06-05

\bibitem{shannon1948}
Shannon, C.E.: A mathematical theory of communication. Bell System Technical Journal  \textbf{27}(3),  379--423 (1948)

\bibitem{shieh2024assessing}
Shieh, A., Tran, B., He, G., Kumar, M., Freed, J.A., Majety, P.: Assessing chatgpt 4.0’s test performance and clinical diagnostic accuracy on usmle step 2 ck and clinical case reports. Scientific Reports  \textbf{14}(1), ~9330 (2024)

\bibitem{sukeda2025japanese}
Sukeda, I., Fujii, T., Buma, K., Sasaki, S., Ono, S.: A japanese language model and three new evaluation benchmarks for pharmaceutical nlp. arXiv preprint arXiv:2505.16661  (2025)

\bibitem{takagi2023performance}
Takagi, S., Watari, T., Erabi, A., Sakaguchi, K., et~al.: Performance of gpt-3.5 and gpt-4 on the japanese medical licensing examination: comparison study. JMIR medical education  \textbf{9}(1),  e48002 (2023)

\bibitem{wei2022chain}
Wei, J., Wang, X., Schuurmans, D., Bosma, M., Ichter, B., Xia, F., Chi, E., Le, Q.V., Zhou, D.: Chain-of-thought prompting elicits reasoning in large language models. Advances in Neural Information Processing Systems  \textbf{35},  24824--24837 (2022)

\bibitem{xie2024medtrinity}
Xie, Y., et~al.: Medtrinity-25m: A large-scale multimodal dataset with multigranular annotations for medicine. arXiv preprint arXiv:2408.02900  (2024), \url{https://arxiv.org/abs/2408.02900}, accessed 8 Jun 2025

\end{thebibliography}

\end{document}